\documentclass[A4, 10pt]{article} 

\usepackage{osameet3} 
\newcommand\authormark[1]{\textsuperscript{#1}}
\usepackage{amsmath,amssymb,parskip,setspace,float}
\usepackage[colorlinks=true,bookmarks=false,citecolor=blue,urlcolor=blue]{hyperref}

\begin{document}

\title{Is it worth the effort? Understanding and contextualizing physical metrics in soccer}

\author{Sergio Llana\authormark{1}, Borja Burriel\authormark{1}, Pau Madrero\authormark{1}, and Javier Fernández\authormark{2}}
\begin{doublespace}
\address{\authormark{1}Barça Innovation Hub, \authormark{2}Zelus Analytics\\}
\end{doublespace}

\section{Introduction}

Physical match performance is one of the most studied topics in Sports Science since EPTS devices became a trend in soccer. Despite the vast number of publications, most research has focused on assessing player performance based on isolated metrics such as distance covered, accelerations, or high-intensity runs (HI) (Bradley et al. 2013; Altmann et al. 2021; Ingebrigtsen et al. 2015). In addition, the tactical context tends to be widely simplified and often ignored. For sports scientists and soccer practitioners, the idea that the integration of tactical and qualitative information can be very beneficial to develop a much more in-depth analysis of physical demands does not go unnoticed. However, the lack of spatiotemporal data that allows analyzing individual effort within the collective context has been an enormous barrier for developing this integration between the physical and the tactical. Far on the horizon remains the old question: is it about running more or running better?

While event and tracking data have accelerated soccer analytics development in recent years, we still lack a comprehensive framework for understanding how physical abilities impact global performance. There are still many unanswered questions: How do player roles affect the type of runs players do? Can we find similar players based on their movements between lines? Can we assess whether a high- intensity disruptive run provides goal value? What types of runs and movements can we expect from our next opponent, and what would its impact be? Can we select a starting eleven that maximizes value creation in specific spaces?

Having a comprehensive approach for addressing these and many other related questions would expand the capabilities of soccer analytics to help the primary decision-makers in football clubs. Some examples are:
\begin{enumerate}
\item The head of recruitment would find players with great physical capacity and contribute to value creation, both on-ball, and off-ball.
\item The head coach and technical directors would assess players' versatility based on their movements and tactical fit to either design tactics or track players' evolution in time.
\item The medical department would better understand fatigue by associating lower physical deployment and contribution to value creation.
\item Physical coaches could better customize training sessions and design tailored routines for players' readaptation and return-to-play by classifying players' movements and identifying those with higher intensity.
\end{enumerate}

We present the first framework that gives a deep insight into the link between physical and technical-tactical aspects of soccer. Furthermore, it allows associating physical performance with value generation thanks to a top-down approach:
\begin{enumerate}
\item We start with a high-level view by identifying differences in physical performance between attack and defense phases, both from a collective and individual perspective.
\item Secondly, we contextualize physical indicators employing spatiotemporal features derived from the interaction of the 22 players and the ball. Specifically, we integrate tactical concepts such as dynamic team formations, player roles, attack types, and defense types.
\item Finally, we employ a state-of-the-art expected possession-value model to associate runs with value creation and its impact on the overall likelihood of winning more matches.
\end{enumerate}

This work is structured into two main parts. First, we present the technical details of how we developed the different layers of our top-down approach. Here, we employ an unprecedented data set of tracking data from broadcast, which allows us to assess and compare various teams and players from the main European leagues. Then, in the second part of the work, we present a broad set of practical applications showing new approaches for understanding players and team performance comprehensively.

\section{Estimating physical metrics from tracking data}

In this work, we count on the availability of TV broadcast tracking data, including near 70\% of the 2020/2021 season's matches from the Big-5 European competitions: the English, Spanish, German, Italian, and French domestic leagues. This novel source of spatiotemporal data provides the location of the 22 players and the ball ten times per second. The availability of this comprehensive data set of tracking data is unprecedented. While EPTS-derived physical metrics might be considered more precise since physical signals are captured directly from devices attached to players' bodies, usually, this information is limited to the 11 players belonging to the team that owns the devices.

To obtain a speed signal, we calculate the player's velocity in two consecutive frames and calculate the magnitude at the frame level. Then, we apply a rolling average smoothing to remove the fine-grained variation between frames. In addition, we detect and treat outliers in players' speed. Figure 1 illustrates the frame-by-frame evolution of Messi's speed before and during an on-ball action (link to the video: \href{https://bit.ly/3qYOgC6}{https://bit.ly/3qYOgC6}).

We detect intervals from the speed signal where a player maintains the speed in the same range, similar to those proposed by Pons et al. (2019). A speed interval is defined as a valley when both the preceding and following periods are of a higher speed. The full speed signal is then divided into sections from each valley until the following one. Each section represents a player's run, and its speed is the maximum speed reached at the peak between these two valleys. For example, in Figure 1, Messi's movement is divided into three different efforts, with peaks of 6, 21, and 21 km/h, respectively.

\begin{figure}[htbp]
  \centering 
  \includegraphics[width=15cm]{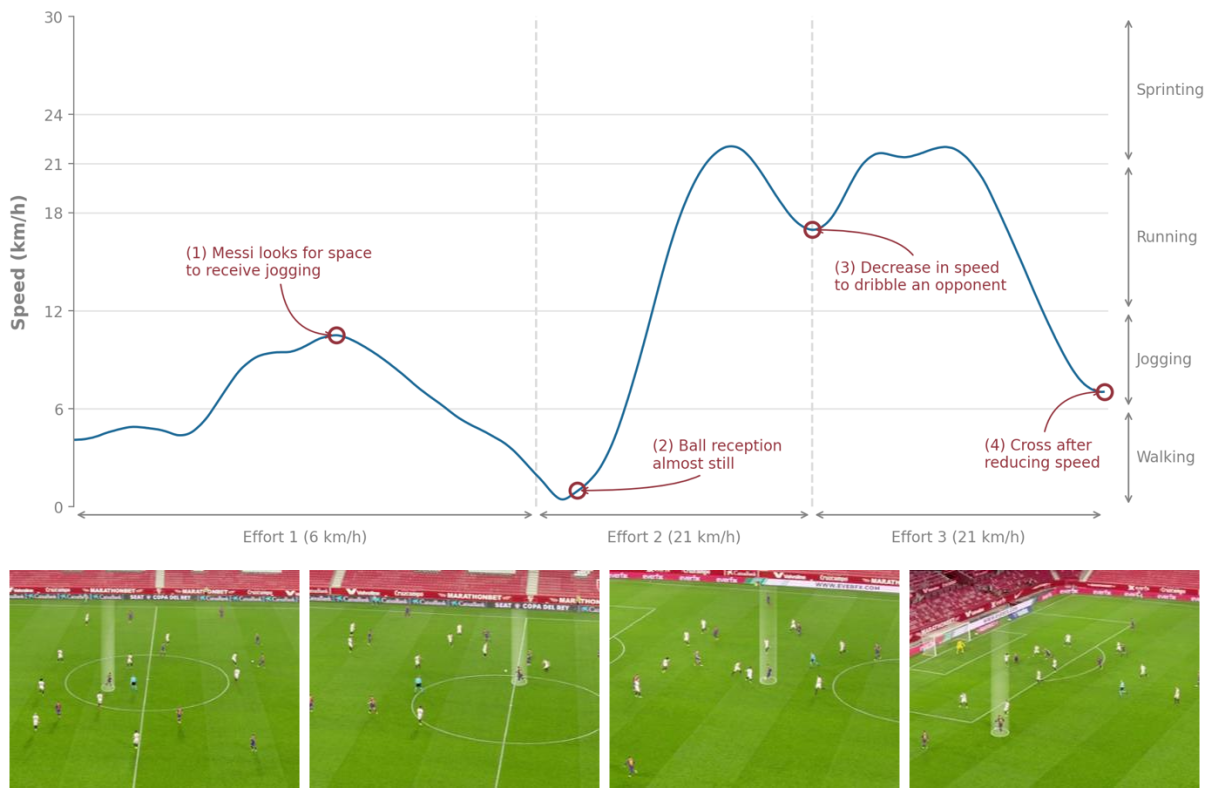}
\caption{Above, the evolution of Messi's speed in a game situation. The first highlighted moment represents a movement to approach (1) the ball. Then, the action is formed on (2) the pass reception, (3) a ball drive with a dribble, and (4) a cross. Below, the video frames of the four highlighted instants.}
\end{figure}

We compute three different metrics to define the physical performance. The most obvious is the distance covered by a player, calculated by adding up each run's distance. Then, we define high-intensity (HI) runs as those exceeding 21 km/h (Castellano et al. 2011; Ade et al. 2016). Given this definition, we are interested in computing the number of runs and the distance covered at a high intensity.

There are three key moments in a player's run used to add the contextual information described in the following section.
\begin{enumerate}
\item The instant when the valley ends. It is used to set the origin location of the run as the player starts to accelerate and increase in speed.
\item The instant when the peak of the run starts. It is the moment used to know if the run is performed within an attack or defense and its derived categories (attack type, defense type, etcetera). It is the moment when the player starts to get to the maximum speed of the run.
\item The instant when the peak of the run ends. It is used to set the destination location of the run as the player starts to decelerate.
\end{enumerate}

\section{A framework for physical performance contextualization}

Soccer is a complex sport and, just as we cannot analyze the parts of a complex system separately, we need to bring together the technical and tactical aspects of the sport with the physical performance. There have been other attempts to integrate these three dimensions of the game in the Sports Science literature, being the approach by Bradley and Ade (2018) the most comprehensive so far.

They associate each player's effort with a tactical concept manually tagged from the video. However, their integrated approach does not consider the team's style, the opponent's block, or the value-added with those efforts, which are essential to contextualizing the physical metrics better.

We have worked on a scalable and automatic way to overcome these issues, which detects contextual tactical concepts by integrating the broadcast tracking data from SkillCorner with StatsBomb's event data. For this paper, we are working with a novel dataset that allows us to have a wide variety of matches from the 2020/2021 season of the Big-5 European competitions (English Premier League, Spanish La Liga, German Bundesliga, French Ligue 1, and Italian Serie A).

\begin{figure}[htbp]
  \centering 
  \includegraphics[width=15cm]{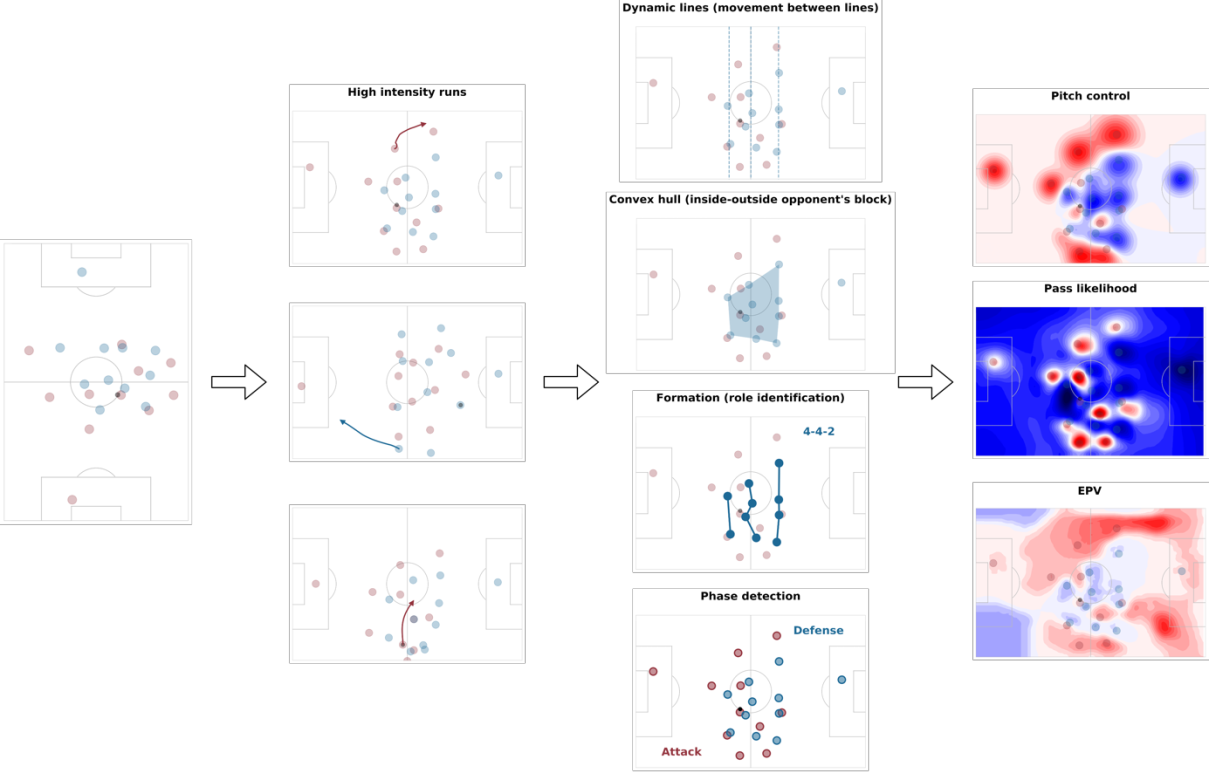}
\caption{The different layers of contextualization of the proposed top-down framework.}
\end{figure}

The top-down framework we propose has different levels of analysis, starting with the physical performance metrics estimated from tracking data with the process described in Chapter 2. Figure 2 shows the layers of contextualization that will be described in the following sections.

\textbf{\subsection{Dynamic lines and team’s block}}

While the absolute location of high-intensity efforts provides a good piece of information on players' movement patterns, contextualizing these efforts according to the opponent's defending block adds a whole new level of tactical insights. Therefore, we integrate the concept of dynamic formation lines introduced by Fernández et al. (2021) to calculate these relative locations.

Representing the defensive structure to compute the relative locations of the events is difficult due to its variability and complexity. However, we assume that players always form three lines, as shown in Figure 3. The centroids of a clustering performed on the defenders' X coordinate model the dynamic lines from tracking.

\begin{figure}[htbp]
  \centering 
  \includegraphics[width=6.5cm]{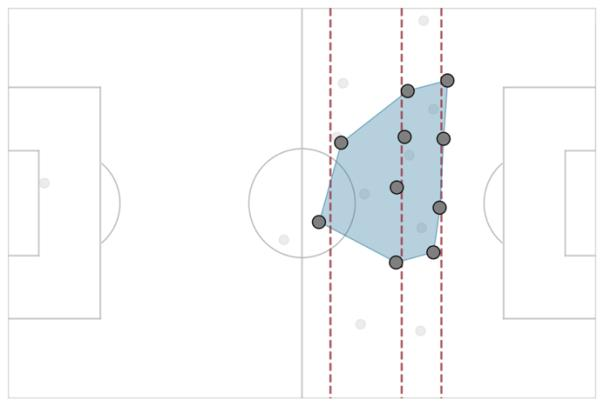}
\caption{Diagram of the dynamic defensive lines (red dashed lines) and the convex hull representing the defensive team's block (blue shaded area).}
\end{figure}

In addition to the dynamic lines, we compute the convex hull of the defender's X and Y coordinates to model the team's block. It helps us distinguish those actions inside the block from those in the flanks. Note that goalkeepers are always ignored to model these two tactical concepts.

To give tactical meaning to each high-intensity run, we have defined three different zones:
\begin{enumerate}
\item Inside: Within the opponent's block and between defensive lines.
\item Wing: Both left and right flanks outside the block but within defensive lines.
\item Back: Everything that is behind the last defensive line.
\end{enumerate}

By combining these relative zones with the runs' origin and destination relative locations, we have defined the types of movements plotted in Figure 4. Note that we have filtered some of the combinations that are not used in the applications of this paper. These categories of HI runs will help us define player profiles, as shown in the second application in Chapter 4. For example, movements inside-to-back or wing-to-back are common traits of deep wingers and strikers.

\begin{figure}[htbp]
  \centering 
  \includegraphics[width=12cm]{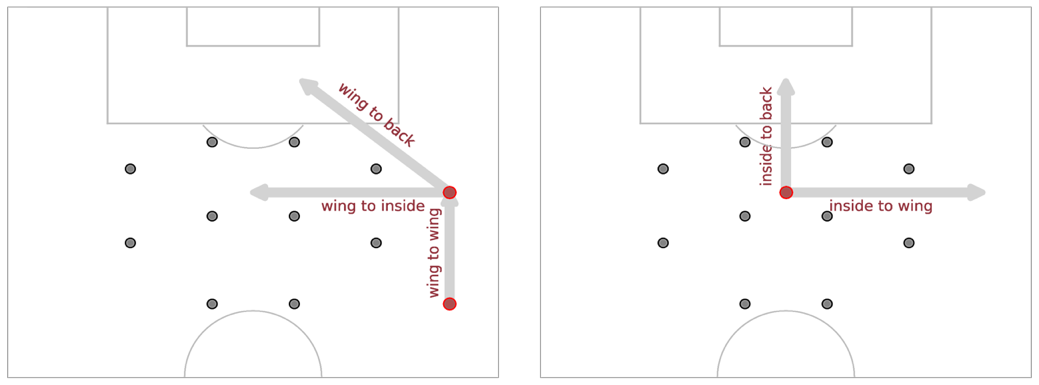}
\caption{Diagram of the types of movements relative to the opponent's block. On the left, those starting from the flank of the block, whereas, on the right, the ones that begin inside the opponent's block.}
\end{figure}

\textbf{\subsection{Possessions, attack types, and defense types}}

Players' responsibilities vary considerably according to who has the ball. The different demands between attacking and defending phases also affect the physical efforts the player has to perform, as studied by Lorenzo-Martínez et al. (2021). Therefore, identifying which team is in possession when a particular run is performed will help add a layer of basic but essential information to contextualize it.

Possessions are detected following a rule-based approach. Three possibilities can cause a possession change: (1) the ball goes out, (2) the referee stops the game, or (3) in-game possession changes, which are fuzzier, and we should take into account those times teams lose the ball and regain it instantly. On the other hand, we apply rule-based algorithms based on player locations, teams' defensive lines, and events at each possession's frame to determine the attack and defense types. We developed and validated these algorithms in coordination with coaches from FC Barcelona to be as close as possible to the way they interpret these tactical concepts.

We identify four possible attack types:
\begin{enumerate}
\item Organized attack: both teams are well-structured with no drastic block movements.
\item Direct play: the segment of the possession after a long vertical pass from the back of the attacking team.
\item Counter-attack: the segment of the possession when both teams travel from one half of the pitch to the other one after a ball recovery.
\item Set-piece: seconds after a set-piece in the opponent's half of the pitch.
\end{enumerate}

The defense types are categorized based on the height of the opponent's block:
\begin{enumerate}
\item High pressure: The block is almost entirely located in the attacking team's half of the pitch, with at least N players in the last third of the pitch.
\item Medium block: The block is distributed in both halves.
\item Low block: The block is entirely on the defending team's half of the pitch.
\end{enumerate}

By automatically detecting these concepts, we can easily define the moment in which a player's run occurs related to the possession of the ball. This way, we can distinguish which runs were performed within an attack, a defense, or out of play (e.g., a central defender comes back from a corner to be ready for the opponent's goal kick.). Finally, we set the attack and defense types regarding the team in possession and the defending team.

\textbf{\subsection{Dynamic team formations and player roles}}

The position or role played within a team substantially impacts the kind of efforts a soccer player will perform during a match (Lorenzo-Martinez et al. 2021). Figure 5 confirms these variations with the distribution of the distance covered at a high intensity per player role when teams were in and out of possession. Note that the differences are especially noticeable when the team is attacking.

\begin{figure}[htbp]
  \centering 
  \includegraphics[width=12cm]{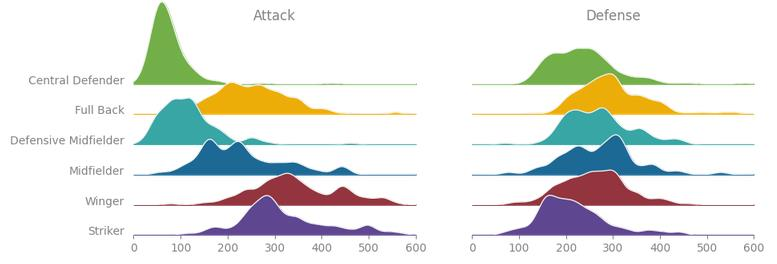}
\caption{Distribution of meters covered at high intensity in attack and defense by each player roles, normalized per 30 minutes of effective playing time in possession (left) and out of possession (right).}
\end{figure}

To identify player roles within a match, we first need to know their team's formation. In addition, this information also impacts players' physical performance, as Bradley et al. (2011) studied. We followed an approach similar to the one presented by Shaw and Glickman (2019), which can be summarized in the following steps: (1) we compute player's mean location relative to their teammates to assemble a formation, then (2) we apply an assignment optimization algorithm between a set of templates that represent the most commonly used systems and the previous players’ mean relative locations. Finally, (3) we assign the formation's positions to each involved player.

In addition, some roles are equivalent among different tactical systems (e.g., a left full-back in a 4-4-2, 4-3-3, and 4-2-3-1), so we choose to simplify them. First, discarding the wing-specific part of the role (e.g., merging left/right-wingers) and then combining similar roles (e.g., full-backs and wing-backs). 

The resulting player roles are central defenders, full backs, defensive midfielders, midfielders, wingers, and strikers. Note that while these roles correspond to the standard set of positions in soccer, these are calculated for every second in the match, allowing for match-related variations rather than relying on global and hand-label positions.

\textbf{\subsection{Linking with Expected Possession Value}}

In the latest years, the estimation of player's actions value has become a trend in soccer analytics research. It is a new dimension that allows turning descriptive analysis into prescriptive by providing qualitative insights to practitioners. In the same way that we estimate the value added by a player's passes, we would like to know which players are running better, adding more value to their team's possessions with their off-ball runs.

There are different approaches in the literature to estimate the long-term probability that a possession ends in a goal. We can categorize them into (1) action-value and (2) possession-value models, depending on how they define the state of the possession. Models such as the ones described by Decross et al. (2020) or the OBV by StatsBomb (2021) are examples of the former, and they use spatiotemporal features to predict the value added by on-ball actions. On the other hand, possession-value models like the ones presented by Rudd (2011), Singh (2019), or Fernández et al. (2019) provide a representation of the game state and estimate the expected value at the end of the possession by integrating over all the possible paths the possession can take from that state. In this work, we use Fernandez et al. (2021) EPV framework, which is built on top of tracking data and considers the impact of both observed and potential actions, providing a rich source of information to assess players' off-ball contribution.

Employing the Expected Possession Value framework, we want to understand the players' runs impact in creating value. In particular, we want to associate the EPV gain between the beginning and end of each HI run to measure the increase in the team's goal probability after the player's effort. The EPV of the attacking team is set at two different moments: (I) when the player starts to increase the speed (the starting valley of the effort ends) and (II) two seconds after the peak of the effort ends. After each high-intensity run ends, we compute the value added to the team's possession with these two values.

We then perform an ordinary least-squares linear regression to obtain the influence of each player's HI runs on their team's EPV increase as shown in equation (1),

\begin{eqnarray}
EPV_{added} \sim \beta_0 + \beta_1 * angle + \beta_2 * distance + \sum_p \beta_p * E_p
\end{eqnarray}

where $angle$ and $distance$ are computed respectively to the opponent's goal from the initial location of the run. $E_p$ represents the player making the high-intensity run. Note that we are considering the player's role per match so that the same player with two different roles will have different coefficients per role.

With this modeling approach, we are essentially estimating a player's average contribution to the possession value when he performs a high-intensity run while controlling for the angle and distance effects.

\textbf{\subsection{Aggregation and normalization}}

Finally, to create a player and a team profile, we need to aggregate per-match metrics for a defined period of time, which is the 2020/2021 season in our case. The most common way of normalization is per match (I.e., 90 minutes), but the truth is that some games have a higher ball-in-play (also known as effective) time than others, with differences of even 20 minutes in total. Therefore, to avoid biasing the results and favoring teams with matches with higher effective time, we choose to do the following normalization:
\begin{enumerate}
\item Those metrics that do refer to a specific possession phase will be normalized per 60 effective minutes, very close to the mean of ball-in-play per match.
\item Metrics referring to events and actions that happen either in attack or in defense will be normalized per 30 minutes of effective time in and out of possession, respectively. Note that for the following applications, we are filtering those players who did not play a minimum of 450 minutes, so that we remove potential outliers due to the lack of minutes.
\end{enumerate}

Note that for the following applications, we are filtering those players who did not play a minimum of 450 minutes, so that we remove potential outliers due to the lack of minutes.

\section{Applications}

Supported by this framework, we now focus on answering a series of commonly asked questions that will serve as practical applications to analyze players' and team behaviors.

\textbf{\subsection{The higher the percentage of possession, the less we run?}}

Pep Guardiola once said: "Without the ball you have to run, but with the ball you have to stay in position and let the ball run.". While coaches often refer to concepts related to running, not all the metrics are interpreted the same way. Some might refer to the total distance traveled, while others might emphasize the more demanding efforts. So, to give a more practical sense to this analysis we will split the initial question into two different questions:
\begin{enumerate}
\item When comparing in and out of possession phases, do teams travel more distance in total?
\item Do teams travel longer distances in attack phases at a high intensity than in defense phases? And does this vary according to the possession percentage?
\end{enumerate}

To answer the first question, we need to control for the effect of the possession percentage. Note that if we sum the total distance covered of a team with higher ball possession, they will look like they run longer in attack because they spend more time in this game phase. Therefore, we decide to normalize these metrics per 30 minutes of effective playing time in and out of possession, respectively. Figure 6 shows the distribution of the distance covered by teams in attack and defense phases, normalized as explained before, and we observe that teams tend to run more when out of possession, confirming Guardiola's idea.

In addition, we also show that the distribution of the distance covered at a high intensity follows the same trend. From now on, we limit our analysis to evaluating performance in high-intensity running since it allows us to differentiate the players' behaviors more clearly, as they are less confused with insignificant movements. At the same time, the FC Barcelona coaches who 
contributed to this study consider that high-intensity runs represent the moments of maximum effort and are the most significant for understanding the player's physical capacity.

\begin{figure}[htbp]
  \centering 
  \includegraphics[width=8cm]{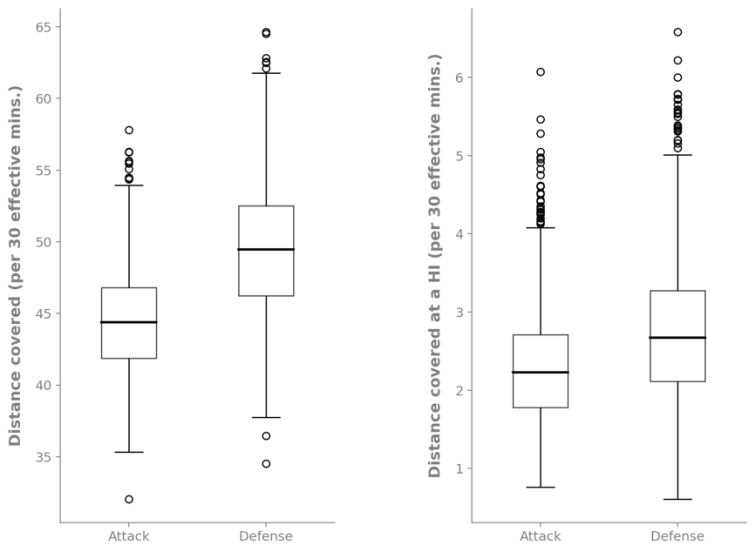}
\caption{Distribution of the distance covered in attack and defense, both in total (on the left) and at a high intensity (on the right). All the metrics are normalized per 30 minutes of effective playing time.}
\end{figure}

Regarding the second question, Lorenzo-Martinez et al. (2021) showed that teams with very high possession percentages run less, both in total and at a high intensity. In our case, we want to evaluate these differences further and isolate them between the attack and defense phases, understanding how team style affects.

Figure 7 shows the teams' normalized distances covered at a high intensity in and out of possession from various European competitions in the 2020/2021 season. The red line separates the teams that run more in defense (above the line) from those that run more when attacking. If we focus on the size of the circles, which represents the mean percentage of possession, we can interpret that teams with a higher ball possession percentage tend to travel more distance at high intensity per minute in defense than in attack.

But is there more correlation between physical and tactical variables? We have carried out a principal component analysis (PCA) to assess the relationships between tactical and physical indicators in teams' style. Figure 8 presents a biplot comparing the teams from the English, German and Spanish leagues in the 2020/2021 season, according to the two principal components, which explain 87.4\% of the total variance. The circle´s color represent the Expected Goals (xG) differential of each team in the mentioned season (accounting exclusively for the games available in the dataset). Note that the translation between team acronyms and the full names can be found in Appendix 1.

\begin{figure}[htbp]
  \centering 
  \includegraphics[width=9.5cm]{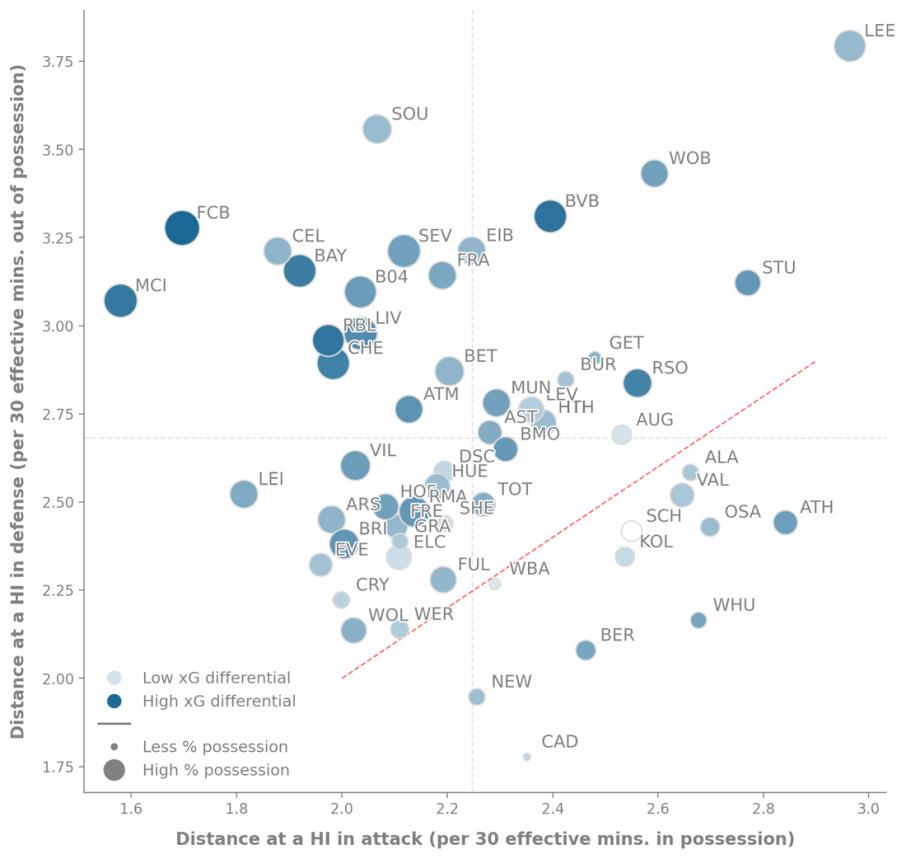}
\caption{The distance at HI covered in attack (X-axis) and defense (Y-axis) normalized per 30 minutes of effective playing time in and out of possession, respectively. The circle's color represents the differential of Expected Goals, whereas the size represents the percentage of ball possession. The red line represents the values whose distance at high-intensity is equal in attack and defense.}
\end{figure}

By analyzing the relationships between variables in the plot, we extract the following conclusions from the teams' offensive and defensive styles:
\begin{enumerate}
\item As previously mentioned, the possession percentage is significantly negatively correlated (-0.78) with running more in defense than in attack.
\item Teams with more possession have a high tendency to have an associative playing style (i.e., less percentage of direct play), avoiding actions with a higher risk of losing the ball.
\item Similarly, teams with more possession tend to have a defensive style based on high press (correlation of 0.81), trying to recover the ball as soon as possible.
\item A higher percentage of direct play, which means more counterattacks, transitions, and long balls, is somewhat correlated (0.55) with more distance at a HI covered in attacks.
\item Finally, more tendency to press higher is significantly correlated (0.78) with running more at HI in defense, as teams with this characteristic will do shorter but more intense efforts.
\end{enumerate}

\begin{figure}[htbp]
  \centering 
  \includegraphics[width=10cm]{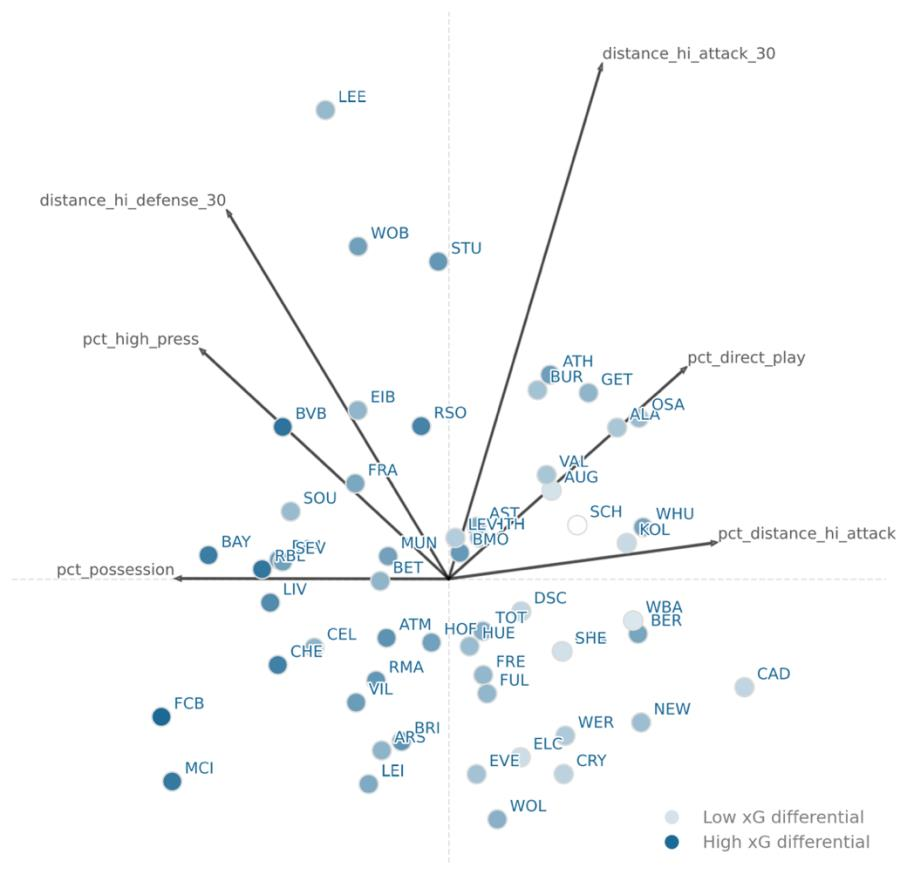}
\caption{Biplot of teams in the two first Principal Components of a PCA with a mix of physical metrics (HI distance covered) and tactical metrics (team's style). All metrics are normalized per 30 minutes of effective playing time in or out of possession, depending on if they refer to the attack or defense phases. The circle's color represents the team differential of Expected Goals.}
\end{figure}

The biplot also lets us grasp the similarity between teams regarding the mix of physical and tactical variables that we have selected. For example, Leeds United shows a striking tendency to run long distances at a high intensity in attack and defense, and only VfL Wolfsburg and VfB Stuttgart are similar to them. In addition, the English team tends to press high. Secondly, teams with higher ball possession percentages such as FC Barcelona, Manchester City, or Bayern Munich are associated with high pressures and low direct play, playing mostly an associative style. These teams are related to a higher ratio of distance covered at a HI in defense.

Finally, teams like Cadiz, Newcastle, or Werder Bremen are related with low percentage possession, based on direct play, and they tend to defend retreating in their own half of the pitch. These teams tend to make more high-intensity efforts in attack. Finally, we can see that the xGoals differential, which we use as a proxy for the team's strength, has no apparent correlation with running more; it depends more on the team style.

\textbf{\subsection{Do all players run the same way?}}

When we want to define the characteristics of a player, the first two dimensions we focus on are the technical and tactical ones. We can approach the former by quantifying the events such as passes, dribbles, drives, or shots and complement it with qualitative information such as the expected value-added. This type of analysis focuses on understanding the purpose of the player within the team's way of playing and other aspects such as the positions they tend to play or their behaviors in the different phases of play. However, a player's profile is not complete without the physical dimension.

Do some players mostly run at a high intensity when interacting with the ball? Does a player tend to do disruptive runs to open up spaces for other teammates? Such questions remained unanswered to this date. In addition, quantifying the average distance covered by the player or the number of efforts at a high intensity is not enough. For example, we will find players that distribute the 10 kilometers they run per match differently, and their behaviors are significantly different with respect to the ball.

Figure 9 demonstrates that players in different roles perform on-ball actions at very different speeds. For example, Central Defenders tend to do more oDistributions of the percentage of on-ball actions each role tends to do at the different speed categories.n-ball actions than the rest of the roles at very low speeds (< 6km/h, walking) and fewer actions sprinting. Still, there is a wide spread of Central Defenders jogging as some can double the percentage of others. On the other hand, more attack-driven roles such as Wingers and Strikers tend to do more actions at very high speeds (> 21 km/h, sprinting), but there is also wide variability in those roles and speed ranges.

\begin{figure}[htbp]
  \centering 
  \includegraphics[width=15cm]{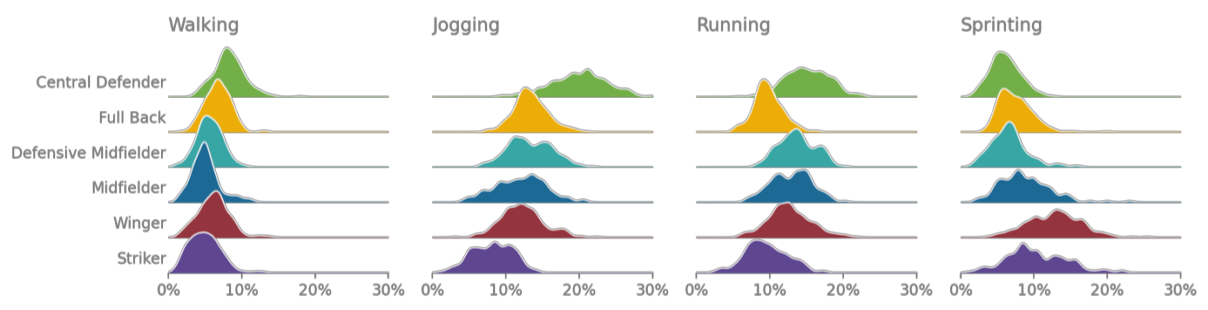}
\caption{Distributions of the percentage of on-ball actions each role tends to do at the different speed categories.}
\end{figure}

However, the proportion of high-intensity efforts spent in on-ball actions is only a proportion of the total HI runs that a player does in attack phases. Figure 10 shows the differences in some of the English, Spanish, and German wingers in 2020/2021. Players such as Messi, Grealish, or Mbappé are wingers who save their energy to those times they have a great chance of getting the ball. In addition, as the circle's size indicates, players such as Saint-Maximin and Adama Traoré stand out for carrying the ball at high speed considerably more often than other players. On the other hand, Cheryshev or Jon Morcillo are players who participate less with the ball but cover more distance at a HI.

The following figure also shows that, on average, from 60\% to 80\% of the winger's HI efforts in the attack phase do not involve direct on-ball participation in the possession, highlighting the importance of assessing off-ball performance in soccer. Unfortunately, this type of off-ball behavior went unnoticed for data-driven analysis of players until now. With the framework we present in this paper, we can quantify the number of times a player does the defined types of off-ball movements described in Chapter 4. More importantly, we can say how unique this number is compared to the other players with the same role in the Big-5 leagues.

\begin{figure}[htbp]
  \centering 
  \includegraphics[width=10cm]{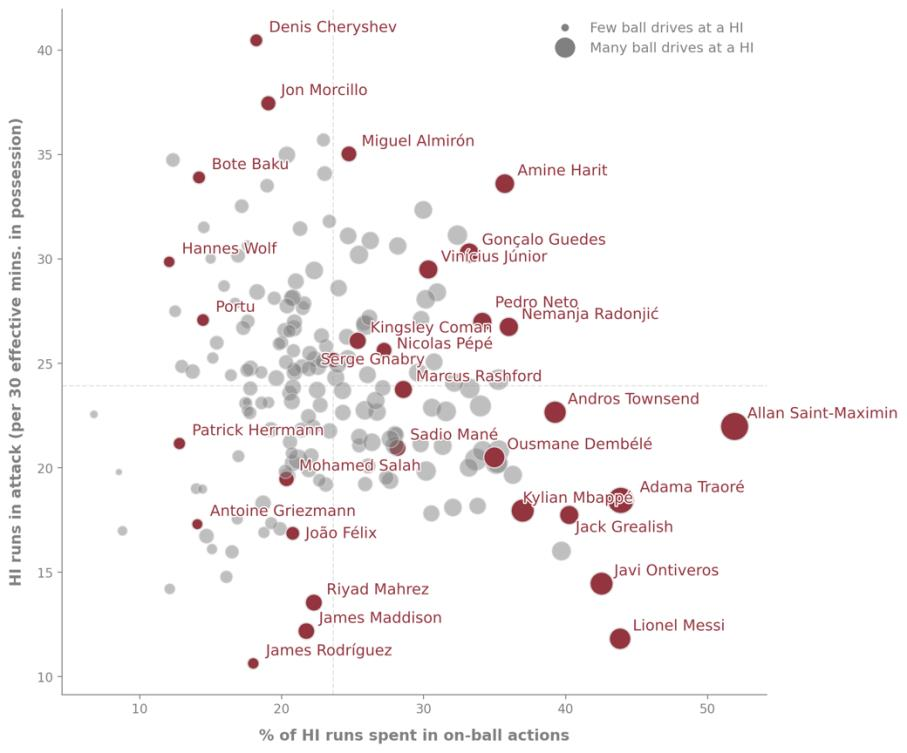}
\caption{Percentage of winger's high-intensity efforts in the attack phase that are performed to participate directly with the ball. The circle's size represents the exact percentage but driving the ball at least 3 seconds. We have the number of HI efforts per 30 minutes of effective playing time in possession in the Y-axis.}
\end{figure}

Figure 11 shows off-ball efforts at a high intensity of Serge Gnabry in Bayern Munich's matches in the 2020/2021 season. Note that the number along each arrow represents the frequency the player did a specific type of HI run per 30 minutes of effective time in possession of the ball. In addition, the color is relative to how rare that frequency is compared to the rest of the wingers (with more than 450 minutes played in that role) of the Big-5 European leagues. The German right-footed player is an all-purpose winger who can play in both the left and right flanks. Thanks to this kind of off-ball behavior analysis, we can now unravel that he does completely different movements when he plays in his natural wing, unlike when he did so as an inverted winger playing on the left flank.

The plots on the top of Figure 11 show that when he plays as an inverted winger, he tends to do high-intensity runs deep behind the back of the defenders (3.0 times per 30 mins of effective time in possession). In contrast, he does not go inside the opponent's block that often. On the other hand, in those matches that he played most of the time as a traditional winger in his natural flank (plots below), he does the opposite behavior with less disruptive runs to the back and running more to the half-spaces inside the block (3.8 times) or staying in the wing (5.3 times). Note from the plots on the right that whenever he starts HI efforts from inside the opponent's block, the behavior is pretty similar regardless of his position within Bayern Munich's tactical system.

This kind of information is precious for a scouting department that seeks to identify the type of winger a player is without watching hours of footage to grasp these insights. And this same pattern is adaptable for the rest of the player roles, creating player profiles that allow us to understand the most common movements at a high intensity of a player. This new dimension of analysis complements tactical-technical studies of frequency and value-added with on-ball actions.

\begin{figure}[htbp]
  \centering 
  \includegraphics[width=12cm]{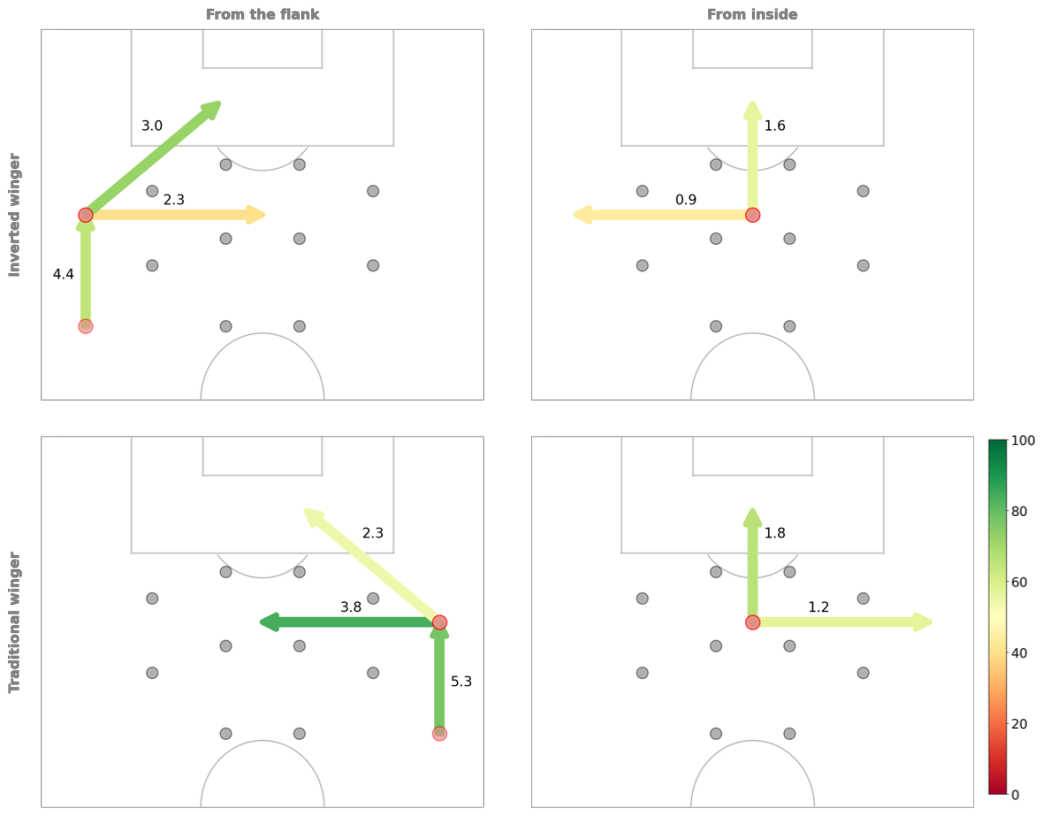}
\caption{Serge Gnabry's frequency of HI runs per match as an inverted winger (above) and a traditional winger (below). The arrow's origin and destination represent the initial and end locations of the effort relative to the opponent's block. The number is normalized per 30 effective minutes in possession of the ball. Arrow's color is relative to the player's percentile compared to what the wingers in the Big-5 European leagues.}
\end{figure}

\textbf{\subsection{Picking up the starting lineup}}

Following the same path and going a step further, we group the normalized frequencies of off-ball run types in all team players. This way, we could answer questions like these: Is the next opponent a deep time from the inside areas, or will their wingers provide most of the depth? Is the opponent symmetric, or do they present different behaviors from the left and right flanks?

We will focus on the 2020/2021 Champions League's winner starting lineup as an example. We have aggregated the frequency of each type of HI runs for all the players that form the 3-4-2-1 used by Chelsea´s coach Thomas Tuchel on that match. Note that the numbers are the full-season averages per player and not the actual number of the final. But how would Chelsea have been less deep if Giroud had played as a striker instead of Werner? Do they keep the same style regardless of the players in the lineup? To answer these questions, we compare the final match´s lineup with an alternative lineup where we substituted the three forwards and the wing-backs.

Observing the differences in the plots of Figure 12, we can state that Chelsea would maintain its style. However, there are two noticeable differences in the HI run distribution.

\begin{figure}[htbp]
  \centering 
  \includegraphics[width=11.5cm]{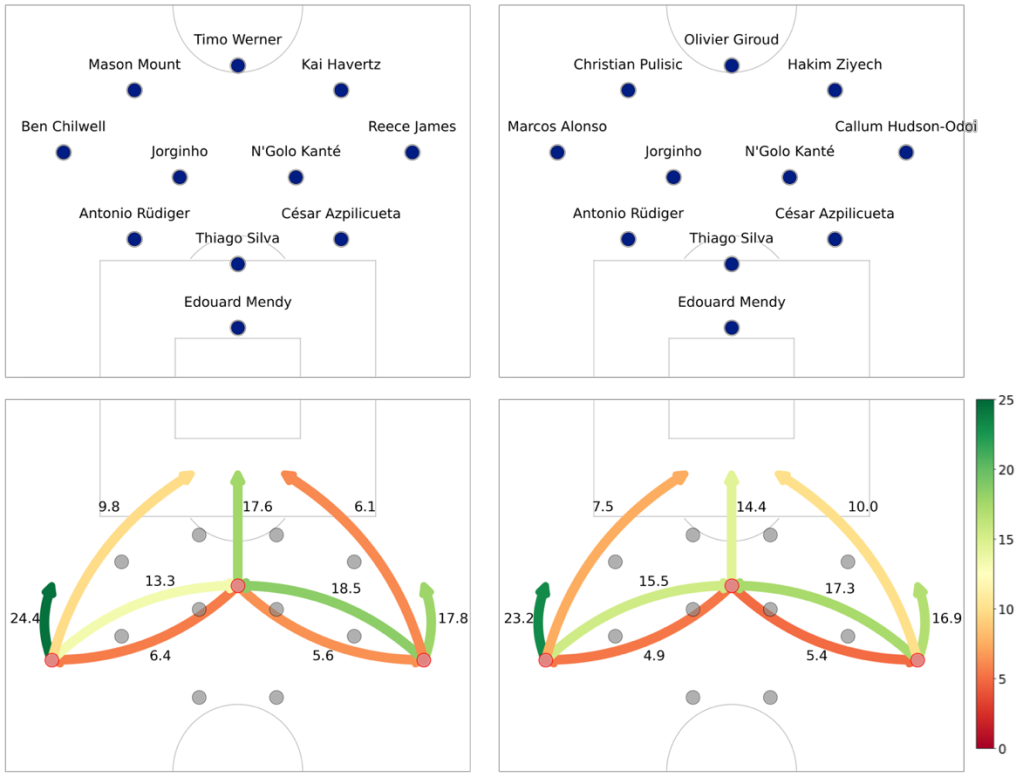}
\caption{Aggregation of the average frequency of the HI runs of all players in two different lineups
normalized per 30 minutes of effective playing time in possession of the ball.}
\end{figure}

Removing Werner from the lineup would decrease the number of disruptive runs per 30 effective minutes of possession both from the inside (from 17.6 to 14.4 times) and from the left wing (from 9.8 to 7.5). In fact, in Havertz's goal in the UCL's final, two inside-to-back runs make the goal chance possible, the former by Werner to drag the central defender and the former by Havertz to face the goalkeeper and score (link to the video of the goal: \href{https://bit.ly/3FPrRuW}{https://bit.ly/3FPrRuW}). Secondly, the team's off-ball runs would be more intense from the right flank than from the left. Maybe because Hudson-Odoi's tends to look for the defender's back.

\textbf{\subsection{How fast does our striker have to be to be valuable?}}

Section 4.2 focused on categorizing the off-ball runs to create offensive profiles of players. But do players get in touch with the ball at the same speed? And what is more important, do they add equal value at low and high intensities? Those questions must be answered too to distinguish different types of players within the same role.

Before focusing on the value-added, we must first understand if players depend more on the high-intensity to receive the ball. We will use a different concept from the standard approach when referring to on-ball actions. Instead of only considering the start and end of passes, crosses, shots, etcetera, we consider the time window between a players’ reception of the ball and the last action he performs within that ball control. Following this, the full extent of Messi’s ball possession shown in Figure 1 would be considered a single action, providing a more nuanced definition involving ball carries.In that path, Figure 13 shows the percentage of times strikers in the Big-5 European competitions were moving at the different speed categories, 2 seconds before receiving the ball. In the background, we plot the 90\% confidence interval. In addition, we highlight five well-known strikers from the top 5 European leagues.

If we focus on Messi, we can observe he is outside the confidence interval in three out of the four speed categories. We see that he is not the classic striker since he tends to receive the ball while walking much more frequently than the rest. On the other hand, it is much less common for him to depend on high-intensity to participate in ball possession. An opposite example is Timo Werner, who tends to receive the ball fewer times at very low intensity than most strikers, but he belongs to the top 5\% of players at high intensities. Finally, we observe that Dybala leans towards receiving after jogging or running compared to the rest of the strikers.

\begin{figure}[htbp]
  \centering 
  \includegraphics[width=10cm]{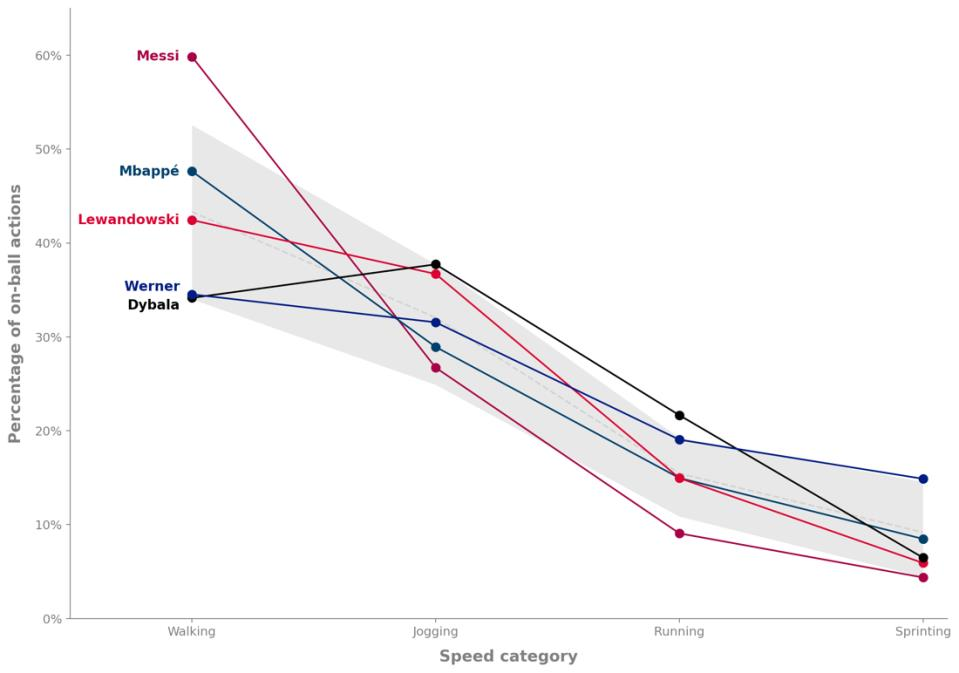}
\caption{Percentage of times that a striker directly participates with the ball at each speed category. The speed is recorded 2 seconds before the ball reception. The gray shaded area represents the 90\% confidence interval among the players with that same role.}
\end{figure}

Such information can greatly benefit a physical coach, who could use these player trends to prepare individualized drills that replicate the specific demands that a player will face on game day. But can they reinforce those types of actions in which the player is significantly decisive? That is when the EPV becomes handy. Figure 14 shows the total EPV added of six strikers from the main European leagues in the different speed categories. Note that EPV added measures the difference in EPV at the end and beginning of each action.

Generally, we observe that strikers generate significant increases in the possession value at higher speeds. We have highlighted two players who could be considered relatively similar: Messi and Dybala. Even though Messi usually generates more value than Dybala with his actions, they both tend to achieve comparable accumulated value-added for the depicted speeds. On the contrary, we detect players like Mbappé or Adama Traoré who add more value with their actions at high intensities.

The examples below and many more extracted from the presented framework are beneficial for both coaches and physical trainers when designing training sessions. Having detailed information on the movements and speeds a player tends to interact with the ball is key, but measuring how much a player influences the team's performance is a game-changer.

\begin{figure}[htbp]
  \centering 
  \includegraphics[width=10cm]{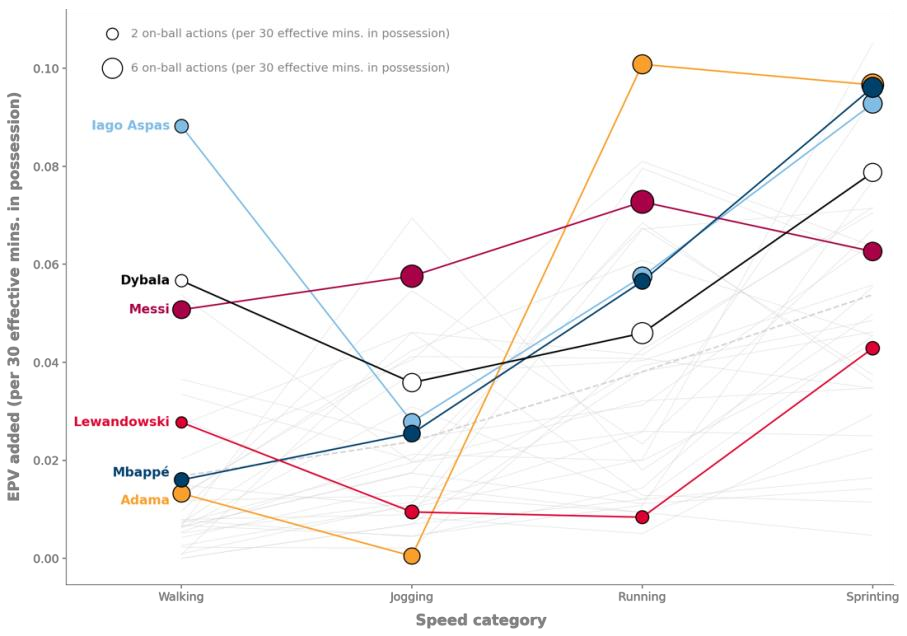}
\caption{EPV added by five strikers with on-ball actions at the different speed categories. The circles' size represents the number of actions the player participates in at each speed category. Both variables are normalized per 30 minutes of effective time in possession. The gray lines in the background represent other strikers from the main European leagues.}
\end{figure}

\textbf{\subsection{What is the impact of high-intensity runs in increasing goal value?}}

The previous section introduced a variable to measure the quality of the on-ball actions, segregating the value by the different speed categories. However, we have not yet discussed adding a qualitative component to all the HI runs a player performs in possession, both on- and off-ball.

To capture the influence of players' HI efforts on the variance of the team's probability of scoring a goal long-term (and at the same time lowering the probability of conceding a goal), we will use the player's coefficients from the linear regression presented in the equation defined in Section 3.4. Both plots in Figure 15 show this new metric's values from Real Madrid and FC Barcelona's players (on the top) and Liverpool and Manchester City's (on the bottom). The metric is then compared to each player's distance covered at HI. Note that some players appear more than once in the plot because we consider those player roles separately if they played at least 450 minutes per role.

We observe how players like Jordi Alba, Nabil Keïta, or Frenkie De Jong (both as a midfielder and as a Central Defender) tend to cover fewer distances at a HI, but they have a strong influence on their team's EPV when they do so. On the other hand, we have previously seen in Section 4.4 that Messi does not tend to receive at a high intensity very often; however, he has an excellent ability for taking advantage of this kind of situation.

\begin{figure}[htbp]
  \centering 
  \includegraphics[width=11.5cm]{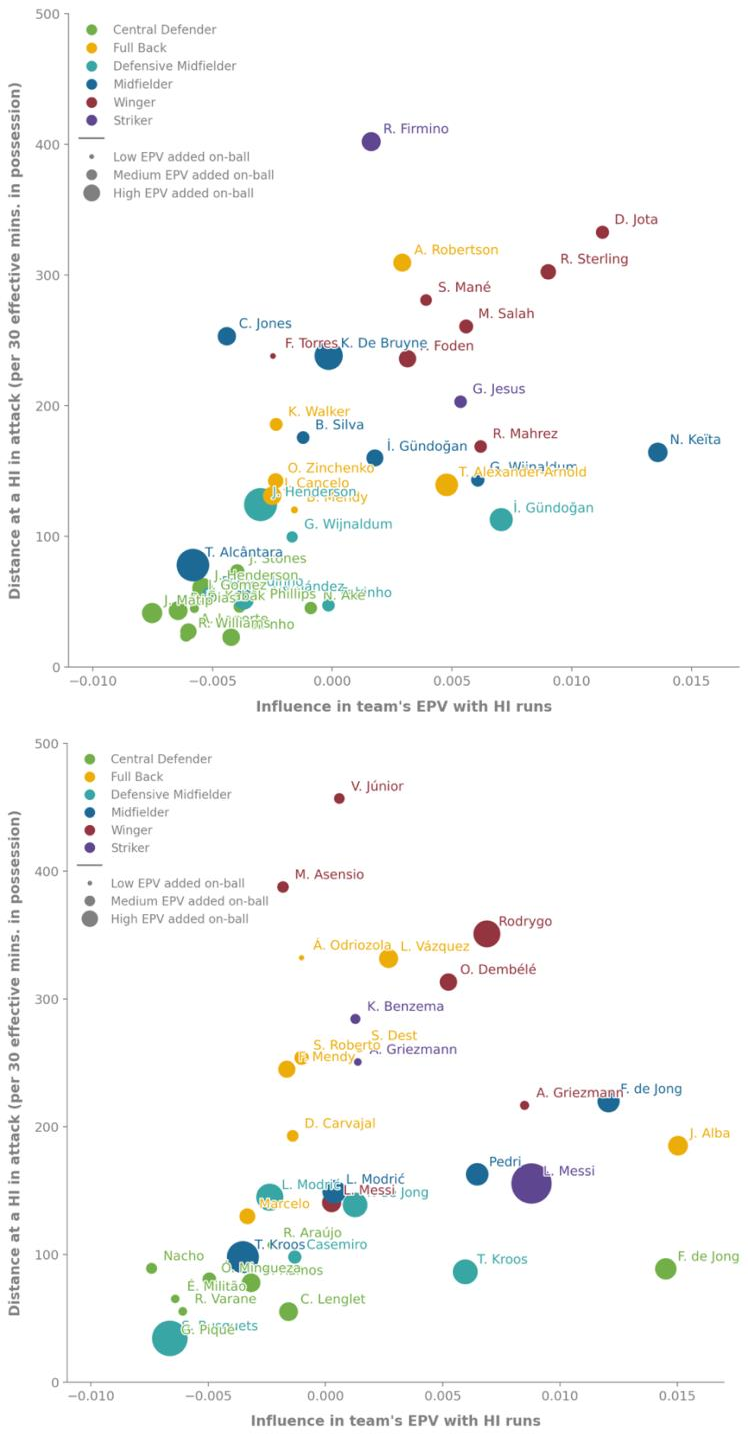}
\caption{Influence of players to their team's EPV after a HI run, extracted from the weights of a linear regression model. Each color is associated with a different player role, and the circle's size represents the on-ball EPV added normalized per 30 minutes of effective playing time in possession.}
\end{figure}

Furthermore, Sergio Busquets and Thiago Alcantara show an influence below average with high-intensity runs compared to all the positions, but a very high on-ball EPV added (demonstrated by their circle's size). This behavior is a clear representation of positional midfielders. On the other hand, players such as Diogo Jota or Raheem Sterling represent a type of attacking midfielder that contributes highly to the team's EPV through their high-intensity runs.

\textbf{\subsection{Do teams maintain high-intensity efforts throughout the entire match?}}

Until now, we have focused on the contextualization of the physical indicators through tactical variables. However, there is a fundamental contextual variable in every team sport: time. In this study, we will not delve into the time factor in-depth; however, we want to do a first examination of how physical effort varies throughout the time of the game.

\begin{figure}[H]
  \centering 
  \includegraphics[width=12cm]{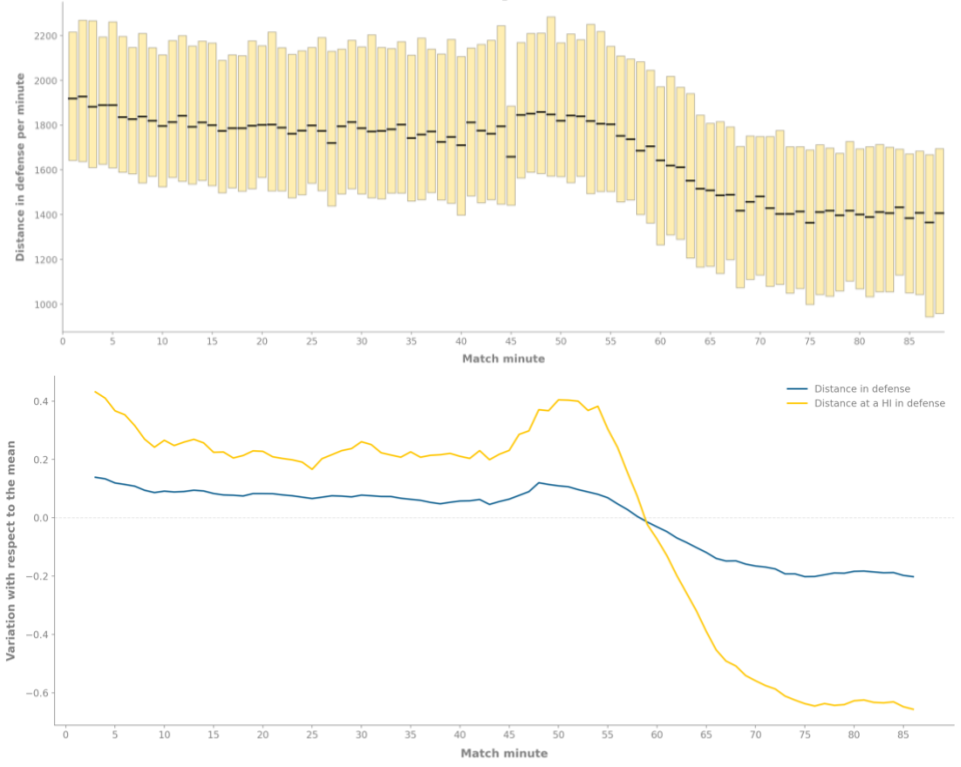}
\caption{Above, the distribution of the distance covered by teams in defense per match minute. Below, the variation with respect to the mean of the distance in defense at a high intensity (in orange), compared to the variation of the global distance (in blue). Both curves are smoothed by applying a rolling average of 5 minutes.}
\end{figure}

Figure 16 shows the evolution of the distance covered in defense in each minute of a match. Note that the metrics are normalized per 60 seconds of effective playing time in out of possession. The plot above presents the distribution of the distance covered at each minute of the game. In contrast, in the plot below, we have computed the variation with respect to the mean of the distance (in blue) and the distance at a high intensity (in orange) in each minute.

We can see a considerable decrease in both the global and the high-intensity distance in defense from minute 65 onwards, but the latter's reduction is much steeper. The reasons for this can be multifactorial, such as the goal differential or fatigue, which despite being a complex concept, we are sure determines performance. Further analysis on this path would help determine the optimal moment to make a substitution or change tactics by putting players on the pitch who exploit the areas that require more defensive effort at the end of the game.

\begin{doublespace}
\section{Discussion}
\end{doublespace}

The presented top-down framework proposes a novel approach for (1) estimating physical indicators from tracking data, (2) contextualizing each player's run to understand better the purpose and circumstances in which it is done, (3) adding a new dimension to the creation of player and team profiles which were often ignored by physical indicators, and finally (4) assessing the value-added by off-ball high-intensity runs by linking with a possession-value model.

In addition to the study's novelty, the framework allows answering practical questions from very different profiles in a soccer club:
\begin{enumerate}
\item First, starting from the ones closest to the grassroots, players could benefit from data-driven feedback and explanations that help them understand the impact of their efforts on their team's performance to boost player development.
\item Secondly, the most direct beneficiaries are coaches and analysts, who can benefit from new tactical insights adding a new dimension to understand better the next opponent and its players merging physic behaviors with tactic information.
\item Then, a scouting department would profit from an improved version of player knowledge, identifying the off-ball profile, which is crucial to understand the player as a whole.
\item Finally, physical coaches and readaptation physiotherapists could also benefit. The former could design individualized drills based on the demands of each player, whereas the latter could understand the type of high-intensity efforts a player tends to do in-match to tailor the exercises in the return-to-play phase fittingly.
\end{enumerate}

In this work, we argued that the simple aggregation of physical indicators such as distance traveled, accelerations, or high intensity runs provide an incomplete and many times insufficient picture of players' actual physical skills regarding their contribution to the game. In the end, soccer is about scoring and preventing goals, but there's been little effort so far in literature to integrate tactical context and value creation with players' physical efforts.

From an individual perspective, we showed how combining fundamental tactical concepts with players' runs provides a richer context for building a player's movement profile and identifying what types of movements occur more often and which create more value. This type of analysis moves research one step forward towards the so-desired individualization of training drills and the understanding of the physical demands of a match so these can be translated into training (and a long-standing goal in sports science).

From a team-level tactical perspective, the proposed contextualization allows coaches to understand
which players would maximize a given strategy, such as exploiting spaces in the sidelines or opening
the field to overload the midfield.

This paper opens a new research path with potential derived studies that could address subjects such as analyzing the optimal moment for the coach to make the substitutions or estimating players' fatigue from a tactical perspective, to identify those zones in the opponent's defense that can be exploited once the match is advanced. In addition, a comprehensive understanding of fatigue's effect on players' decision-making and technical performance would directly improve player development.

\begin{doublespace}
\end{doublespace}

\end{document}